\documentclass[runningheads]{llncs}

\usepackage{iciap}

\usepackage{iciapabbrv}

\usepackage{graphicx}
\usepackage{booktabs}
\usepackage[bottom]{footmisc}
\usepackage{subcaption}
\usepackage{array}
\usepackage[accsupp]{axessibility}
\usepackage{hyperref}

\usepackage{orcidlink}

\newcommand\nnfootnote[1]{%
  \begin{NoHyper}
  \renewcommand\thefootnote{}\footnote{#1}%
  \addtocounter{footnote}{-1}%
  \end{NoHyper}
}

\begin{document}

\title{Benchmarking Content-Based Puzzle Solvers on Corrupted Jigsaw Puzzles} 

\author{Richard Dirauf\inst{1,2}\orcidlink{0000-0003-0820-8589} \and
Florian Wolz\inst{2}\orcidlink{0000-0001-8734-978X} \and
Dario Zanca\inst{1}\orcidlink{0000-0001-5886-0597} \and
Björn Eskofier\inst{1}\orcidlink{0000-0002-0417-0336}}

\authorrunning{R.~Dirauf \etal}

\institute{Machine Learning and Data Analytics Lab, Department Artificial Intelligence in Biomedical Engineering, Friedrich-Alexander-Universität Erlangen-Nürnberg (FAU), 91052 Erlangen, Germany \\
\email{\{richard.dirauf,dario.zanca,bjoern.eskofier\}@fau.de}\\
\and
Siemens Healthineers AG, 91301 Forchheim, Germany}

\maketitle

\nnfootnote{This preprint has not undergone peer review or any post-submission improvements or corrections. The Version of Record of this contribution is published in \\
\textit{\textbf{ICIAP 2025, Lecture Notes in Computer Science 16167 (2026) pp 286-298}}, \\
and is available online at \url{https://doi.org/10.1007/978-3-032-10185-3_23}}

\begin{abstract}

    Content-based puzzle solvers have been extensively studied, demonstrating significant progress in computational techniques. However, their evaluation often lacks realistic challenges crucial for real-world applications, such as the reassembly of fragmented artefacts or shredded documents. In this work, we investigate the robustness of State-Of-The-Art content-based puzzle solvers introducing three types of jigsaw puzzle corruptions: missing pieces, eroded edges, and eroded contents. Evaluating both heuristic and deep learning-based solvers, we analyse their ability to handle these corruptions and identify key limitations. Our results show that solvers developed for standard puzzles have a rapid decline in performance if more pieces are corrupted. However, deep learning models can significantly improve their robustness through fine-tuning with augmented data. Notably, the advanced \textit{Positional Diffusion} model adapts particularly well, outperforming its competitors in most experiments. Based on our findings, we highlight promising research directions for enhancing the automated reconstruction of real-world artefacts. 

\end{abstract}

\section{Introduction}
\label{txt:intro}

Real-world combinatorial optimization problems are NP-hard, complex, and time-consuming to solve. In addition to these challenges, automatic solvers must handle noisy data and non-optimal surrogate objectives, making robustness a critical requirement.
One category of these real-world problems includes the reassembly of objects such as shredded documents \cite{marques_reconstructing_2009} or fragmented historical artworks \cite{papaodysseus_contour-shape_2002, pintus_survey_2016}. This challenge is in close relationship to solving jigsaw puzzles as both are defined similarly: Given a set of pieces, the goal is to reconstruct the original image or object, from which the pieces were created \cite{markaki_jigsaw_2023}. Throughout the years, many automated solvers for jigsaw puzzles were developed utilizing both heuristic and deep learning methods \cite{markaki_jigsaw_2023} with the hope of also being applicable to real-world problems.

Despite significant advancements, a recent study \cite{tsesmelis_re-assembling_2024} demonstrates that nearly all existing jigsaw puzzle solvers are unsuitable to work on realistic archaeological fragments and those that can be applied perform poorly. This is because the fragments are eroded, irregularly shaped due to fractures, vary in size, or are missing due to weathering or human-induced damage. Such characteristics of a real-world archaeological puzzle are often not present in the datasets that are used to develop jigsaw puzzle solvers. Therefore, the authors identify a research gap between existing computational methods and the automated reconstruction of real-world artefacts.

To bridge this gap, it is essential to evaluate the robustness of existing solvers by systematically assessing their ability to handle these real-world challenges. Since isolating these characteristics from an archaeological dataset is impossible, a practical approach is to introduce controlled corruptions to standard jigsaw puzzles mimicking the complexities of real-world artefacts. These types of corruptions can be categorized into two groups: The first group involves modifications to the shape of a puzzle piece, such as irregular contours, multiple edges, or varying piece sizes \cite{harel_crossing_2021, harel_pictorial_2024}. The existing approaches to these challenges focus on fitting together the contours of neighbouring pieces, while mostly disregarding the pictorial content of the fragments. 
The second group includes corruptions that alter the visual content of a puzzle, such as erosion of texture and colour, the degradation of the edges of a piece, or the complete absence of multiple pieces. Most content-based jigsaw puzzle solvers are designed to reassemble uncorrupted, square-sized pieces \cite{markaki_jigsaw_2023} and their robustness to such modifications remains largely unexplored.

In this paper, we focus on content-modifying corruptions and systematically benchmark the State-Of-The-Art (SOTA) content-based jigsaw puzzle solvers to investigate their robustness to different types of corruptions. Using the \textit{PuzzleWikiArts} dataset \cite{talon_ganzzle_2022} we introduce three types of corruptions to simulate real-world fragment characteristics. We then evaluate the performance of selected solvers, comparing heuristic methods and deep learning models, to determine their effectiveness and limitations. Furthermore, we explore whether learning-based methods can be adapted to handle corruption more effectively. Based on our structured investigation, we derive suggestions for the most promising research directions for developing automated methods capable of reconstructing historical artefacts.

\section{Related Work}
\label{txt:related-work}

When reviewing existing research, we focus on 2D content-based image puzzles with square-sized, non-rotated pieces, also known as Type 1 jigsaw puzzles \cite{gallagher_jigsaw_2012}, as it is the most frequently studied type in the literature \cite{markaki_jigsaw_2023}. We identify the SOTA puzzle solvers to test them in our benchmark by examining the comparative investigation by Giuliari \etal \cite{giuliari_positional_2024}, which compare three heuristic solvers and four deep learning models, and Son \etal \cite{son_solving_2019}, which compiled the results reported for nine heuristic solvers.

\subsection{Content-Based Heuristic Jigsaw Puzzle Solvers}

Over the years, many heuristic algorithms have been proposed to automatically solve Type 1 jigsaw puzzles. Most of these methods consist of two key components: a metric to compute the pairwise compatibility between neighbouring puzzle pieces and a strategy to assemble the puzzle based on this metric.
One of those pairwise compatibility measures, introduced by Gallagher \cite{gallagher_jigsaw_2012}, is the Mahalanobis Gradient Compatibility (MGC). It evaluates the local gradients at the boundaries of the jigsaw puzzle pieces, based on the assumption that neighbouring pieces should exhibit similar gradient distributions at their edges. Gallagher shows that MGC significantly outperforms previous compatibility metrics which rely on pixel intensity differences at piece boundaries. This is also confirmed by Giuliari \etal \cite{giuliari_positional_2024} and Son \etal \cite{son_solving_2019}. Gallagher combines the MGC with a greedy, tree-based reassembly algorithm to solve different types of jigsaw puzzles.

The pairwise compatibility metric by Paikin and Tal \cite{paikin_solving_2015} assesses the $L1$ norm of the boundary pixels between all pieces as a measurement of dissimilarity. Their deterministic heuristic identifies pairs of pieces with the smallest and most unique differences, as these are likely to be neighbours, and uses them as starting points to greedily reconstruct the original image. Their method achieves high reconstruction accuracies, outperforming Gallagher's solver \cite{son_solving_2019, giuliari_positional_2024}.

Yu \etal \cite{yu_solving_2015} utilize MGC as their pairwise compatibility metric and combine it with a linear program, which exploits all pairwise matches simultaneously. According to the comparison by Son \etal \cite{son_solving_2019}, Yu \etal's approach achieves the highest performance on two out of five jigsaw puzzle datasets and the second-highest performance on the remaining three datasets. 

Overall, heuristic solvers show strong performance on the Type 1 jigsaw puzzle. Their main limitation lies in their pairwise compatibility metrics, which are not universally optimal for all possible variations of image puzzles. Among different approaches, the two most effective compatibility measurements are the MGC by Gallagher and the $L1$ norm by Paikin and Tal. Therefore, we select both solvers for our benchmark and additionally include the heuristic solver by Yu \etal due to its high performance.
We refrain from analysing additional heuristic solvers in our experiments, as they either exhibit significantly lower performance compared to the selected approaches or lack open-source implementations, making it difficult to reproduce their reported results.

\subsection{Content-Based Deep Learning Jigsaw Puzzle Solvers}
 
In recent years, many researchers tried to apply deep learning models to the jigsaw puzzle problem. These methods follow a two-step process: first, visual features are extracted from the individual image pieces by a neural network. In the second step, the original permutation or the absolute positions of the puzzle pieces are predicted based on the processed features. Therefore, deep learning methods have the advantage that they can learn the reassembly task by leveraging both structure- and colour-based features across the entire image \cite{markaki_jigsaw_2023}. However, earlier deep learning approaches were limited to small puzzle sizes or a fixed number of possible input permutations. 

Talon \etal \cite{talon_ganzzle_2022, talon_ganzzle_2025} are the first who tested their model on puzzles up to $12 \times 12$ pieces but their performance is still behind those of the best heuristics. As one of the few papers in this field, they test how different approaches handle missing pieces, Gaussian noise, or missing borders in the puzzle. They report a performance drop for all investigated solvers but do not analyse the results further.

A recent paper by Giuliari \etal \cite{giuliari_positional_2024} introduced \textit{Positional Diffusion}, a method based on Diffusion Probabilistic Models that is designed to solve ordering problems. For image puzzles, the method learns to predict the positions of the pieces in a normalized coordinate system by perturbing their position vectors during the diffusion process. 
The authors report that their method outperforms other solvers, for example, Paikin and Tal's approach, across most of the tested puzzle sizes. Notably, one of the compared deep learning solvers that is also developed by the same authors is a Transformer model that likewise predicts the pieces' coordinates but without the diffusion process.

In summary, research on applying deep learning models to the jigsaw puzzle problem is still limited, and only a few approaches achieve performance comparable to existing heuristic solvers. As a result, we include two deep learning models in our benchmark: the Transformer-based model and \textit{Positional Diffusion}, both by Giuliari \etal.
\footnote{Implementations: \href{https://github.com/vahanhuroyan/PuzzleDemoGCL}{Gallagher, Yu \etal}, \href{https://github.com/ZaydH/sjsu_thesis/tree/master/mixed_bag_solver/paikin_tal_solver}{Paikin and Tal}, \href{https://github.com/IIT-PAVIS/Positional_Diffusion}{\textit{Positional Diffusion}}}

\section{Methodology}
\label{txt:methods}

\begin{figure}[b]
    \centering
    \begin{subfigure}{0.24\textwidth}
        \centering
        \includegraphics[width=\linewidth]{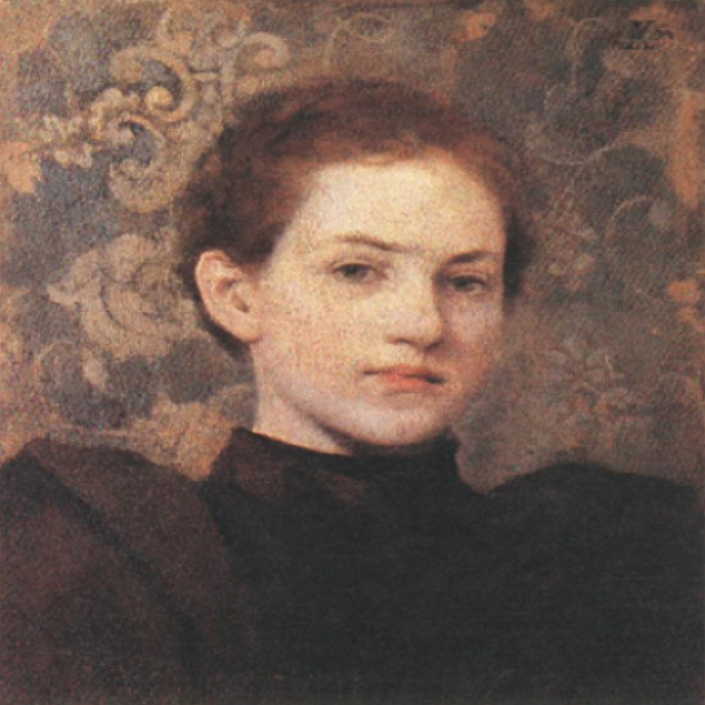}
        \caption{Original Image}
        \label{fig:Example_Image_standard}
    \end{subfigure}
    \begin{subfigure}{0.24\textwidth}
        \centering
        \includegraphics[width=\linewidth]{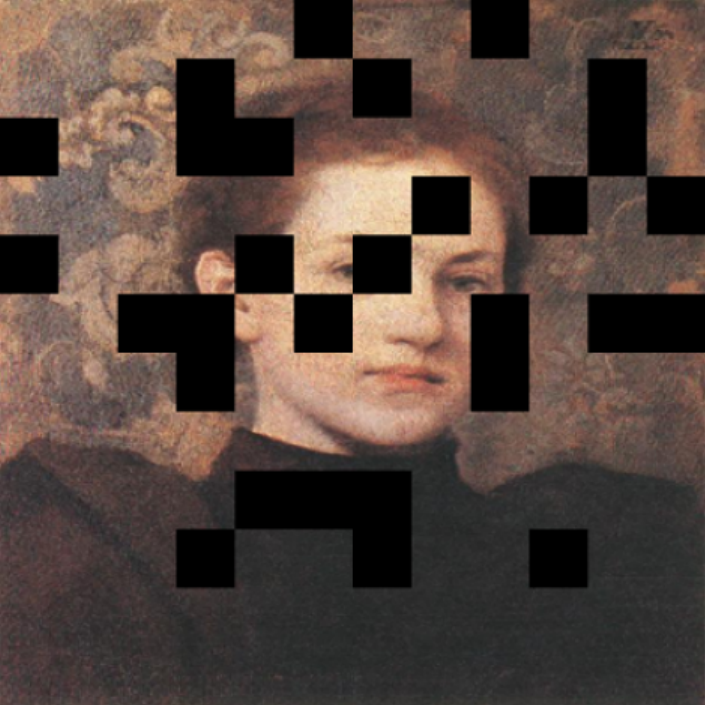}
        \caption{Missing Pieces}
        \label{fig:Example_Image_missing_pieces-20}
    \end{subfigure}
    \begin{subfigure}{0.24\textwidth}
        \centering
        \includegraphics[width=\linewidth]{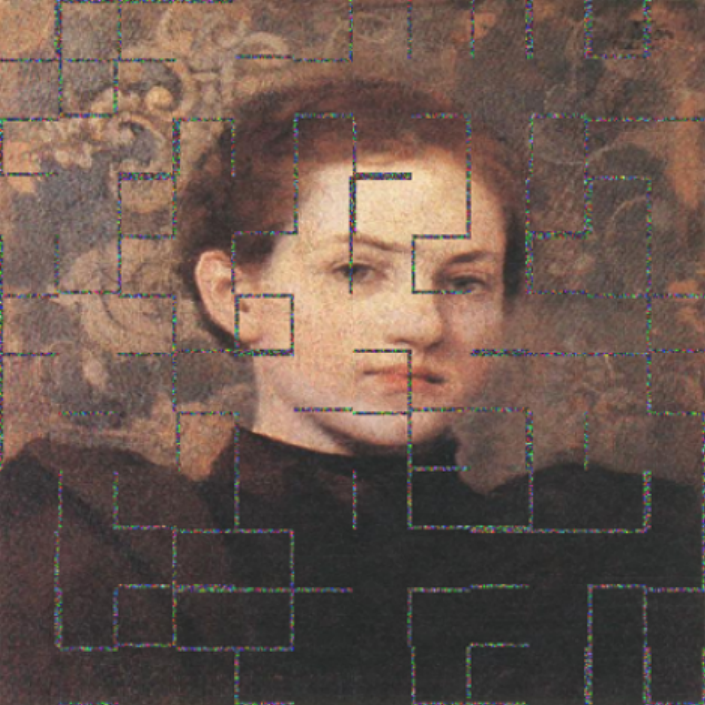}
        \caption{Eroded Edges}
        \label{fig:Example_Image_eroded_edges-30}
    \end{subfigure}
    \begin{subfigure}{0.24\textwidth}
        \centering
        \includegraphics[width=\linewidth]{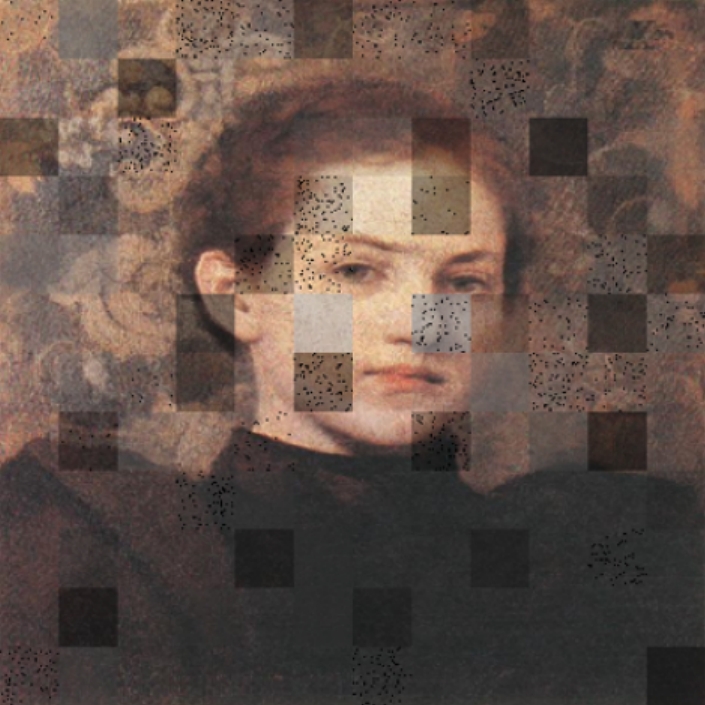}
        \caption{Eroded Contents}
        \label{fig:Example_Image_eroded_contents-20}
    \end{subfigure}
    \caption{Example of a $12 \times 12$ puzzle modified by each of the three corruption types. The original image is taken from the \textit{PuzzleWikiArts} dataset \cite{talon_ganzzle_2022}.}
    \label{fig:overall}
\end{figure}

This chapter presents our approach to benchmark the five selected, SOTA content-based jigsaw puzzle solvers on corrupted puzzles to determine their robustness. To test the performance of the solvers, we are analysing two established metrics \cite{gallagher_jigsaw_2012}. The \textit{Direct Comparison Metric} measures the percentage of pieces correctly placed in their absolute positions within the puzzle. The \textit{Perfect Reconstruction Metric} is the percentage of jigsaw puzzles from a dataset that could be correctly reassembled into the original image. 
For our experiments, we define three types of corruptions inspired by real-world archaeological puzzles and describe their systematic application to unmodified jigsaw puzzles: missing pieces, eroded edges, and eroded contents.

\subsection{Types of Corruption for Jigsaw Puzzles}

\subsubsection{Missing Pieces}

The first type of corruption includes missing pieces. Therefore, a percentage of the total number of pieces is removed from the puzzle and the solvers are tasked with reconstructing the original image as accurately as possible using the remaining pieces. The missing pieces are randomly selected, which can lead to holes in the image, as shown in \cref{fig:Example_Image_missing_pieces-20} if the proportion is low. At higher percentages, solvers may need to assemble disconnected components of the image, further increasing the complexity of the task. In each experiment, the percentage of missing pieces is gradually increased and we limit the experiments to a maximum of 50\%. 
Since all three heuristic solvers do not inherently support missing pieces, the selected pieces are replaced with black image patches of the same shape, with pixel values set to zero. In comparison, it is valid to remove the selected missing pieces from the input for both investigated deep learning models as they predict the absolute position of each puzzle piece. 
To assess the robustness of the puzzle solvers against this type of corruption, we analyse the two performance metrics. Since only the correct placement of the non-missing pieces should be considered, the computation of these metrics is adapted accordingly.

\subsubsection{Eroded Edges}

To simulate eroded edges on square-piece jigsaw puzzles in our benchmark, we adopt the approach of Huroyan \etal \cite{huroyan_solving_2020}, who use this type of corruption to compare three heuristic solvers. Specifically, we randomly select any of the four edges of a puzzle piece with a fixed probability. For each selected edge, the two outermost pixel rows or columns are replaced with new values, which are uniformly sampled from the entire image. An example of this corruption is shown in \cref{fig:Example_Image_eroded_edges-30}.
While Huroyan \etal examine eroded edges with probabilities up to 14\%, we extend this range by gradually increasing the probability up to 50\%. At this maximum level, an average of two edges per piece are affected, severely limiting the puzzle solvers' ability to compare neighbouring edges and reconstruct the original image.

This type of corruption does not precisely replicate the irregularly shaped erosion seen in archaeological fragments. However, it serves as a valid approximation for Type 1 jigsaw puzzles. In real-world scenarios, eroded edges make it difficult to match fragments based on their boundaries. Similarly, our simulated erosion prevents solvers from using affected edges for the reassembly process. As a result, the methods must rely on the remaining uncorrupted edges or the internal content of the pieces to reconstruct the puzzle.

\subsubsection{Eroded Contents}

The third category of corruption includes erosion suffered to the pictorial contents. As the causes and the severity of damage to an artwork differ for each real-world example, it is difficult to simulate all possibilities of decay. Therefore, we focus on four general effects that can impact historical paintings which are exposed to erosion, specifically loss in saturation, contrast and brightness as well as the complete loss of paint through flaking \cite{podany_art_2023}. 

To control the amount of erosion to the contents of a jigsaw puzzle, we introduce the erosion factor $\mathcal{E} \in [0, 100]$ (in \%) which controls the corruption of the pieces in two ways. First, we assume that in a realistic fragmented artwork, each fragment is affected differently by erosion, meaning that some parts could be unharmed while others are severely damaged. $\mathcal{E}$ controls the proportion of pieces that are modified by each of the four erosion effects. This means that a piece in the jigsaw puzzle can be either unmodified or modified by one or a combination of multiple corruption types. Secondly, $\mathcal{E}$ also controls the severity of the individual erosion effect. If a piece is selected for corruption, the proposition of the effect is sampled from a discrete list using a normal distribution centred around $\mathcal{E}$. As a result, the severity of the erosion is generally increased with higher $\mathcal{E}$, while each piece of a puzzle is affected differently by the four effects. A visual example of this type of corruption can be seen in \cref{fig:Example_Image_eroded_contents-20}.

\subsection{Content-Based Jigsaw Puzzle Dataset}

We apply the selected corruptions on the \textit{PuzzleWikiArts} dataset \cite{talon_ganzzle_2022}. It is derived from \textit{WikiArts} \cite{tan_improved_2019} and includes 63 thousand images of paintings, featuring a wide variety of contents and artistic styles. The images are commonly split into puzzles of $6 \times 6$ up to $12 \times 12$ pieces with each piece consisting of $32 \times 32$ pixels. For testing, the four puzzle sizes are uniformly distributed across the test set, resulting in approximately 3125 images per puzzle size.

\subsection{Experimental Design}
\label{txt:experiments}

For all solvers, we use their standard implementations, with adjustments made only to the data loading and evaluation procedures to accommodate the specific types of corruption. 
Before benchmarking the solvers on corrupted puzzles, the SOTA solvers are first tested on standard Type 1 jigsaw puzzles without any corruption. This step is done to reproduce the performance of the four selected approaches as reported in \cite{giuliari_positional_2024}. Additionally, we compare the solvers based on their overall performance across the two evaluation metrics. This serves as a baseline before proceeding with the subsequent experiments, where we investigate the robustness of the jigsaw puzzle solvers towards missing pieces (\textbf{Experiment 1}), eroded edges (\textbf{Experiment 2}) and eroded contents (\textbf{Experiment 3}).

We extend the evaluation of the deep learning solvers by not only measuring the performance of the base models, which were trained on uncorrupted jigsaw puzzles but also by fine-tuning both models through data augmentation. By comparing the resulting performances, we assess how well the deep learning methods can adapt to this type of corruption by modifying only the input data while keeping the solver implementations unchanged.

\section{Results}
\label{txt:results}

\subsection{Baseline: Standard Jigsaw Puzzles}

To establish a baseline for the subsequent experiments, we evaluate all five selected jigsaw puzzle solvers on standard Type 1 puzzles. The results are presented in \cref{tab:base_results_wikiart}. Comparing our results with those of Giuliari \etal \cite{giuliari_positional_2024} on the \textit{Direct Comparison Metric}, we successfully reproduce the performance of three out of four solvers. Only for the solver by Paikin and Tal, we observe a significant divergence, with an average performance gap of 13.85\% compared to the previously reported results. 
The reason for this discrepancy remains unclear, especially since the exact implementation used in the previous study is not specified.

For Type 1 jigsaw puzzles, the best-performing method is the heuristic by Yu \etal, which demonstrates high accuracy in reconstructing various puzzle sizes. In contrast, the other two heuristics show significantly worse performance. Among the deep learning models, the diffusion-based model outperforms the Transformer model, which shows a sharp decline as the puzzle size increases. While \textit{Positional Diffusion} is competitive with the best heuristic methods in the \textit{Direct Comparison Metric}, particularly for smaller puzzles, it struggles to accurately reconstruct larger puzzles. 

\begin{table*}[b]

\caption{Comparison of the solvers' performances in reconstructing standard Type 1 puzzles. The best-performing results are highlighted in \textbf{bold}, second-best results are \underline{underlined}.}
\label{tab:base_results_wikiart}

\centering
\begin{tabular}{@{}l|cccc|cccc@{}}

\toprule
\multicolumn{1}{r|}{\textbf{Metrics}}
    & \multicolumn{4}{c|}{\textbf{\textit{Direct Comparison}}} 
    & \multicolumn{4}{c}{\textbf{\textit{Perfect Reconstruction}}} \\
\midrule
\textbf{Solvers} & \textbf{6x6} & \textbf{8x8} & \textbf{10x10} & \textbf{12x12} 
    & \textbf{6x6} & \textbf{8x8} & \textbf{10x10} & \textbf{12x12}\\ 
\midrule

Gallagher & 86.67 & 85.06 & 81.26 & 79.19 & 81.73 & 78.59 & 75.10 & \underline{71.65} \\
Paikin and Tal & 82.71 & 81.77 & 80.53 & 80.81 & 74.99 & 70.92 & 65.55 & 63.83 \\
Yu \etal & \underline{98.73} & \underline{98.09} & \textbf{96.83} & \textbf{95.63} & \textbf{97.78} & \textbf{95.73} & \textbf{93.25} & \textbf{91.57} \\
\midrule
Transformer & 98.24 & 96.12 & 90.49 & 80.23 & 90.74 & 72.59 & 36.88  & 6.45 \\
\textit{Positional Diffusion} & \textbf{99.14} & \textbf{98.32} & \underline{96.79} & \underline{93.62} & \underline{96.88} & \underline{91.08} & \underline{81.53} & 67.50 \\ 
\bottomrule

\end{tabular}
\end{table*}

\subsection{Experiment 1: Corrupting Jigsaw Puzzles with Missing Pieces}

\begin{figure}[t]
    \centering
    \includegraphics[width=\linewidth]{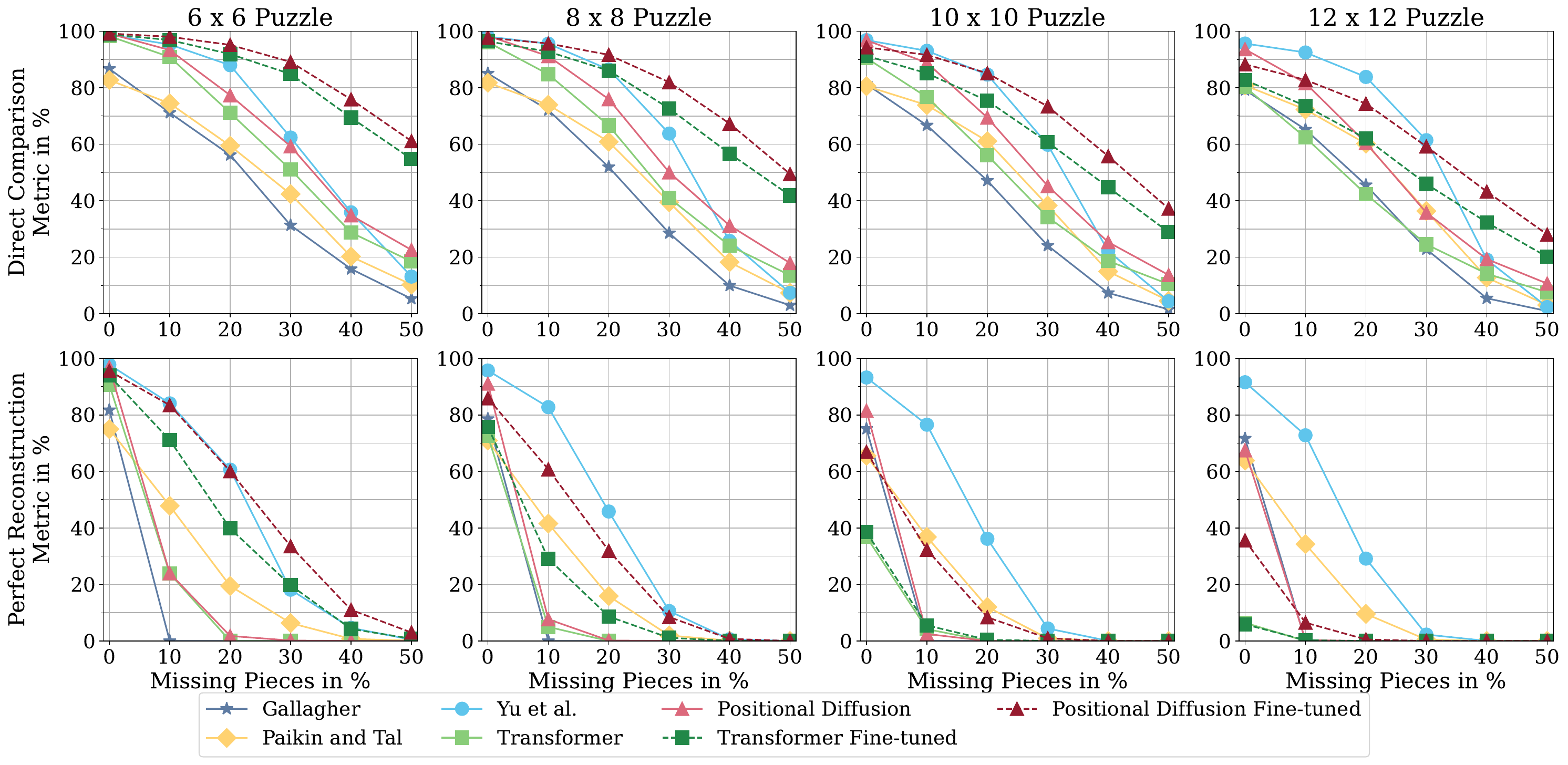}
    \caption{Results of \textbf{Experiment 1} showing the performance of all solvers on puzzles with missing pieces for four different puzzle sizes.}
    \label{fig:Missing_Pieces}
\end{figure}

In general, the performance of all jigsaw puzzle solvers declines as the number of missing pieces increases. This trend is evident in the evaluation metrics across all four puzzle sizes, as shown in \cref{fig:Missing_Pieces}. Among the heuristics, Yu \etal's solver generally achieves the highest performance, followed by the methods of Paikin and Tal and then Gallagher. 
The Transformer and \textit{Positional Diffusion} base models also perform worse than Yu \etal's heuristic. 

Fine-tuning the deep learning models on puzzles with missing pieces significantly improves their performance. Although the Transformer model still lags behind, \textit{Positional Diffusion} outperforms Yu \etal's heuristic on smaller puzzles with up to $8 \times 8$ pieces in the \textit{Direct Comparison Metric}. However, for larger $12 \times 12$ puzzles, Yu \etal's solver remains superior when 30\% or fewer pieces are missing. Additionally, the heuristic reconstructs more jigsaw puzzles perfectly than \textit{Positional Diffusion} for puzzles of size $8 \times 8$ or larger.
The visual example in \cref{tab:example_images_experiment} presents the prediction of the five puzzle solvers on a jigsaw puzzle with 10\% missing pieces. In this example, the three heuristics and \textit{Positional Diffusion} can reconstruct the original image perfectly, while the fine-tuned Transformer model misplaces multiple pieces.

\subsection{Experiment 2: Corrupting Jigsaw Puzzles with Eroded Edges}

\begin{figure}[t]
    \centering
    \includegraphics[width=\linewidth]{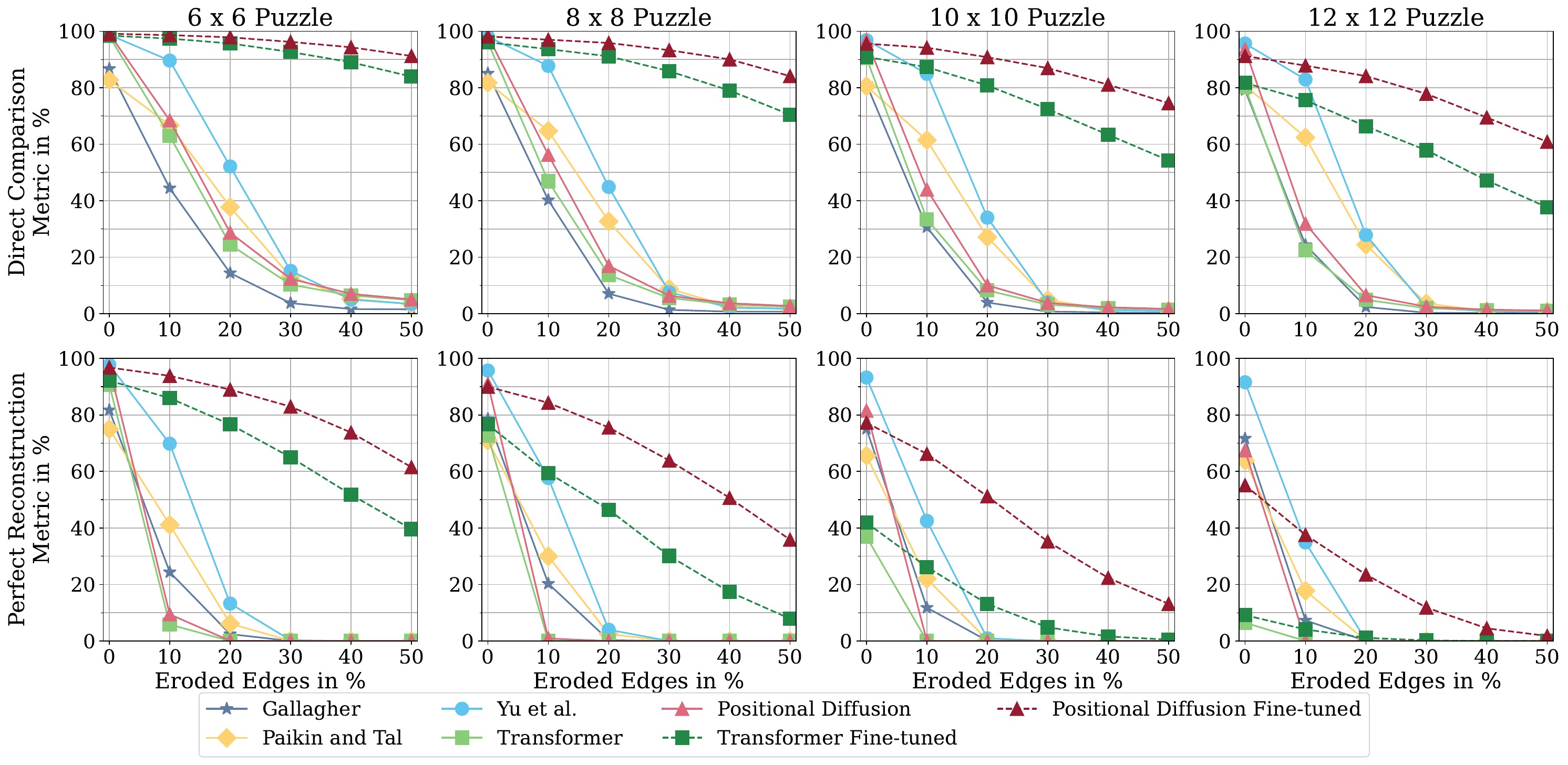}
    \caption{Results of \textbf{Experiment 2} showing the performance of all solvers on puzzles with eroded edges for four different puzzle sizes.}
    \label{fig:Eroded_Edges}
\end{figure}

When jigsaw puzzles are corrupted with eroded edges, the performance of all three heuristics and both deep learning base models declines rapidly, as shown in \cref{fig:Eroded_Edges}. Among these, Yu \etal's solver achieves the highest accuracies, followed by Paikin and Tal’s and Gallagher's heuristic. The \textit{Positional Diffusion} and Transformer base models are on par with Gallagher on the \textit{Direct Comparison Metric}, but perform significantly worse on the other metric. However, as the level of corruption increases to 30\% or more, the differences between the approaches are diminishing and all five methods fail to reconstruct most puzzles.

Fine-tuning the deep learning models on this type of corruption again significantly improves their performance. Notably, \textit{Positional Diffusion} becomes the best-performing method, outperforming both Yu \etal's solver and the Transformer model in nearly all scenarios. However, it still faces challenges on larger puzzles, successfully reconstructing fewer than half of them when the edges are corrupted.
The visual example of \cref{tab:example_images_experiment} confirms the previous findings. There, Gallagher's solver fails to reassemble a quadratic image while the other methods misplace multiple pieces that are severely affected by this type of corruption. The prediction by \textit{Positional Diffusion} is almost perfect.

\subsection{Experiment 3: Corrupting Jigsaw Puzzles with Eroded Contents}

\begin{figure}[t]
    \centering
    \includegraphics[width=\linewidth]{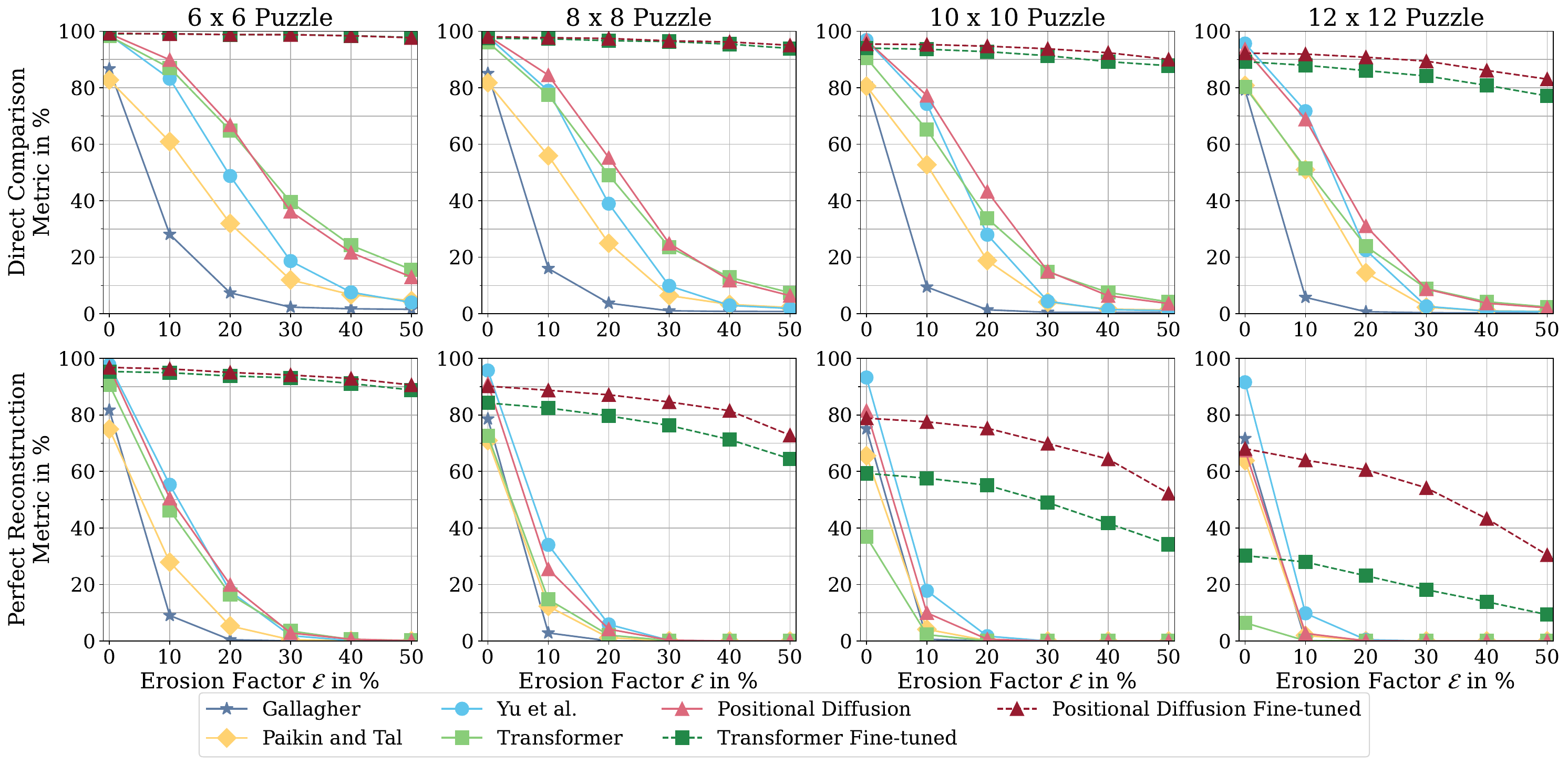}
    \caption{Results of \textbf{Experiment 3} showing the performance of all solvers on puzzles with eroded contents for four different puzzle sizes.}
    \label{fig:Eroded_Contents}
\end{figure}

\Cref{fig:Eroded_Contents} presents the performance of the jigsaw puzzle solvers across different levels of content erosion, controlled by $\mathcal{E}$, for all four puzzle sizes. Among the heuristics, Gallagher's solver performs the worst, followed by Paikin and Tal’s method, with Yu \etal's heuristic achieving the highest accuracies. However, the performance for all three heuristics declines rapidly and they struggle to reassemble most test images when faced with eroded contents. While the base models of the deep learning approaches follow a similar downward trend, their \textit{Direct Comparison Metric} remains higher for smaller puzzles up to $8 \times 8$ pieces.

Fine-tuning the deep learning models on this type of corruption significantly improves their performance. \textit{Positional Diffusion} remains the most effective solver, successfully reconstructing the highest number of puzzles in all scenarios. The Transformer model, while less effective than the diffusion model, still outperforms all other solvers by a considerable margin. 
A visual example is presented in \cref{tab:example_images_experiment} where the heuristics misplace every piece. The reconstructed image by the Transformer model shows more consistent structures while also misplacing many pieces. \textit{Positional Diffusion} almost perfectly assembles the original image.

\begin{table}[t]

\caption{Visual examples of the predictions from the three heuristics and two fine-tuned deep learning solvers on three types of corruptions. Each example is a $12 \times 12$ puzzle from the \textit{PuzzleWikiArts} dataset \cite{talon_ganzzle_2022}. Pieces placed at the wrong absolute position are marked with red dots.}
\label{tab:example_images_experiment}

\centering
\setlength\tabcolsep{1.0pt}
\begin{tabular}{m{0.16\textwidth} | *{5}{m{0.16\textwidth}}}
\toprule

\centering \textbf{Experiment Details}
    & \centering \textbf{Gallagher}
    & \centering \textbf{Paikin and Tal}
    & \centering \textbf{Yu \etal}
    & \centering \textbf{Transformer}
    & \centering \textbf{\textit{Positional Diffusion}} \arraybackslash\\
\midrule

\centering Missing\\ Pieces\\ (10\%)
    & \centering \includegraphics[width=0.16\textwidth]{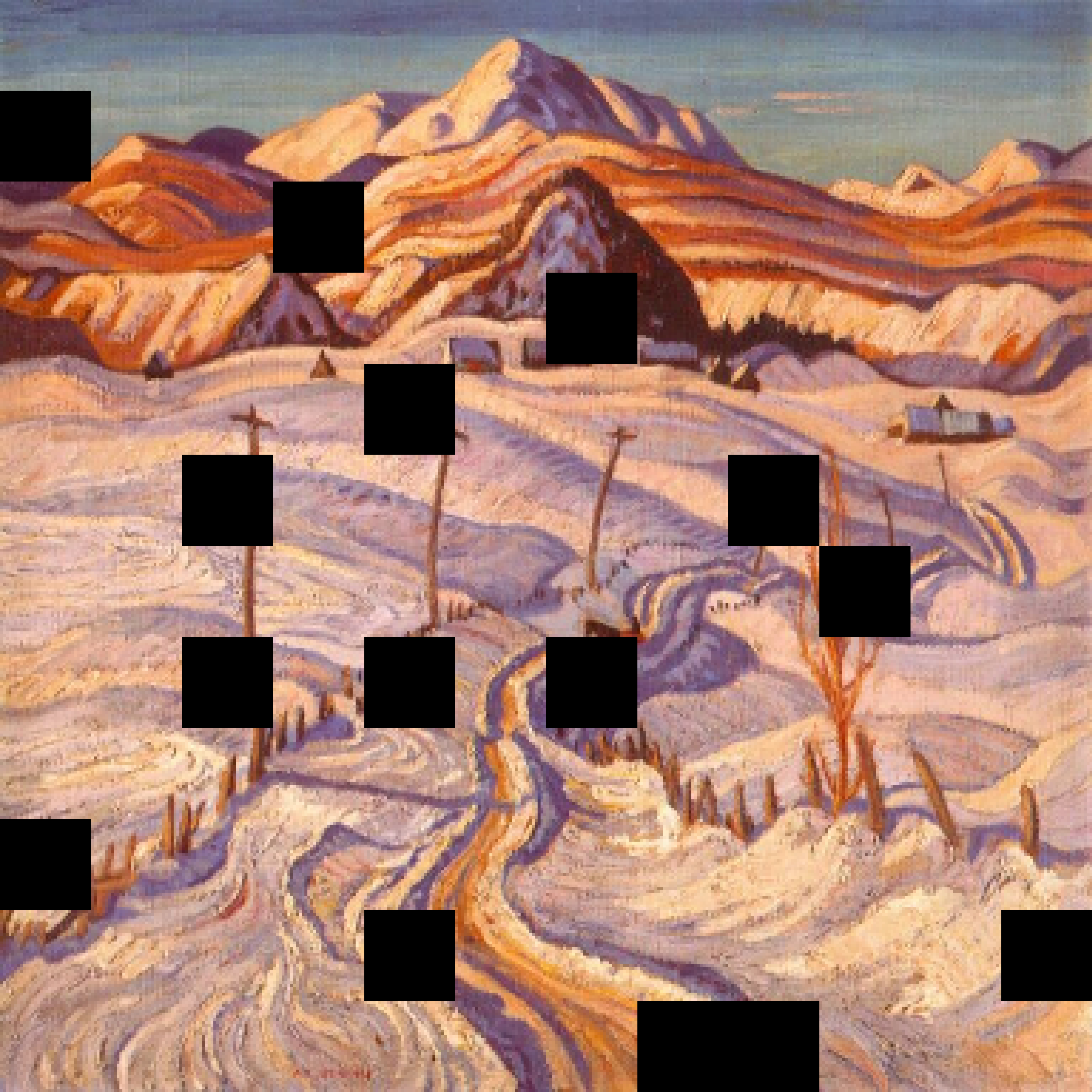}
    & \centering \includegraphics[width=0.16\textwidth]{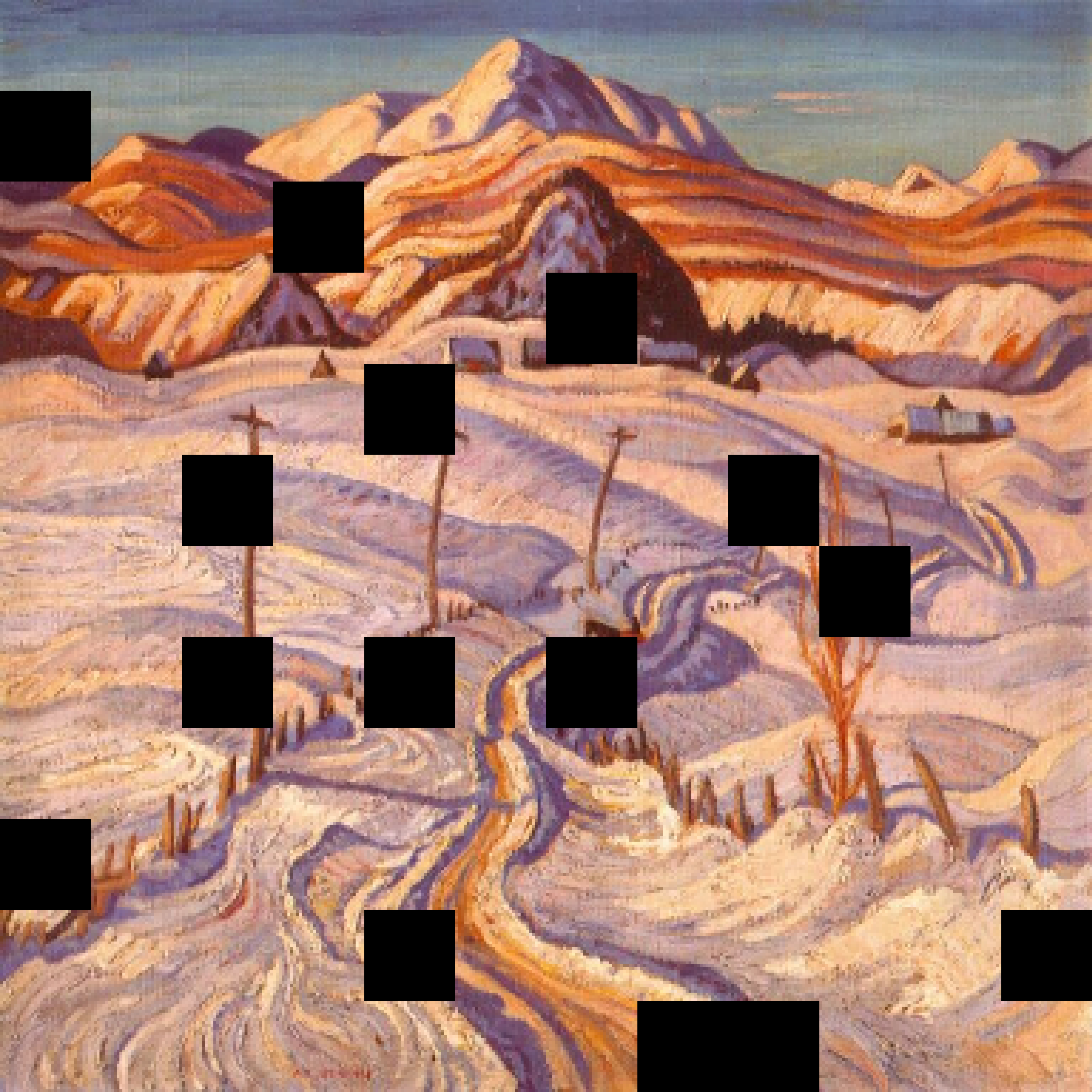}
    & \centering \includegraphics[width=0.16\textwidth]{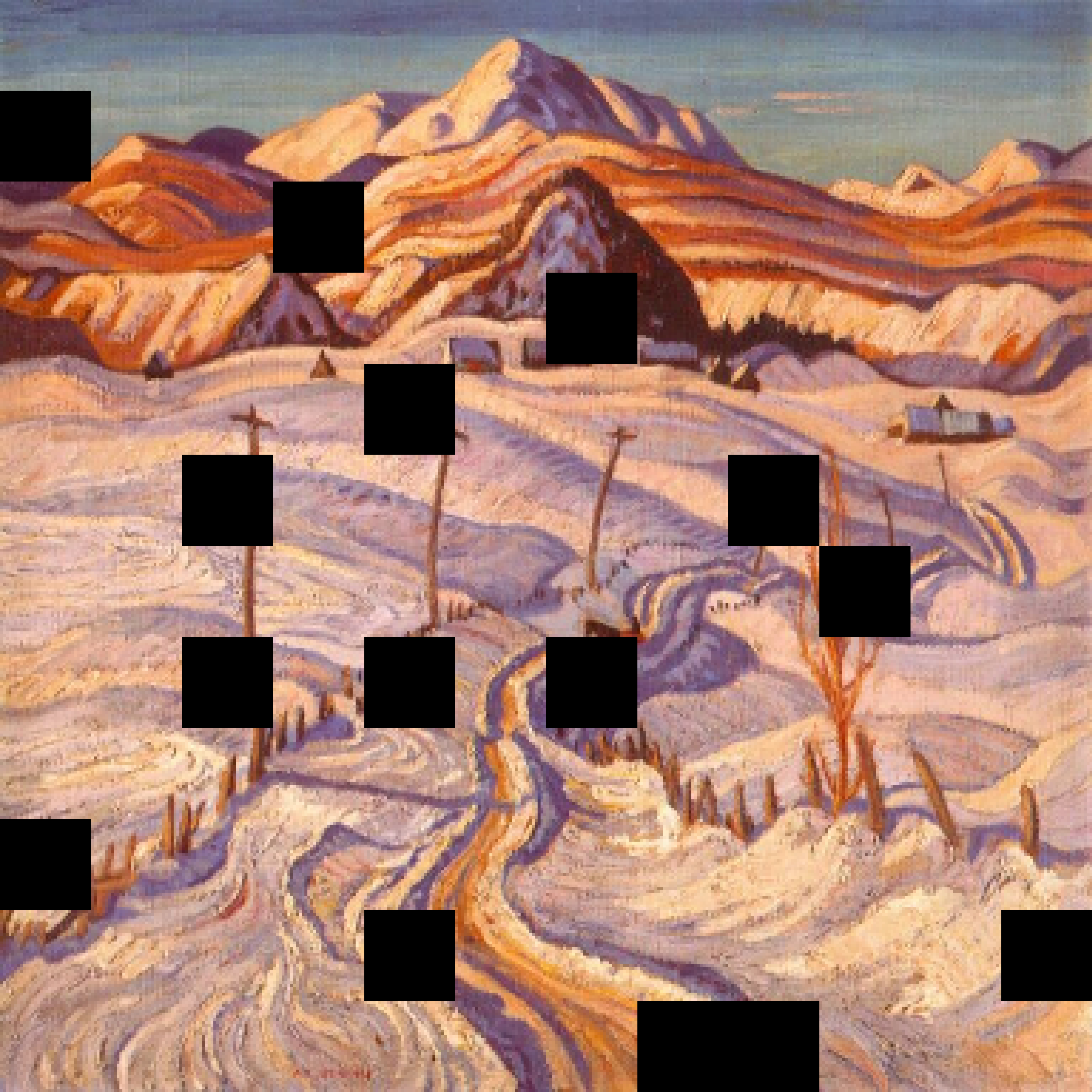}
    & \centering \includegraphics[width=0.16\textwidth]{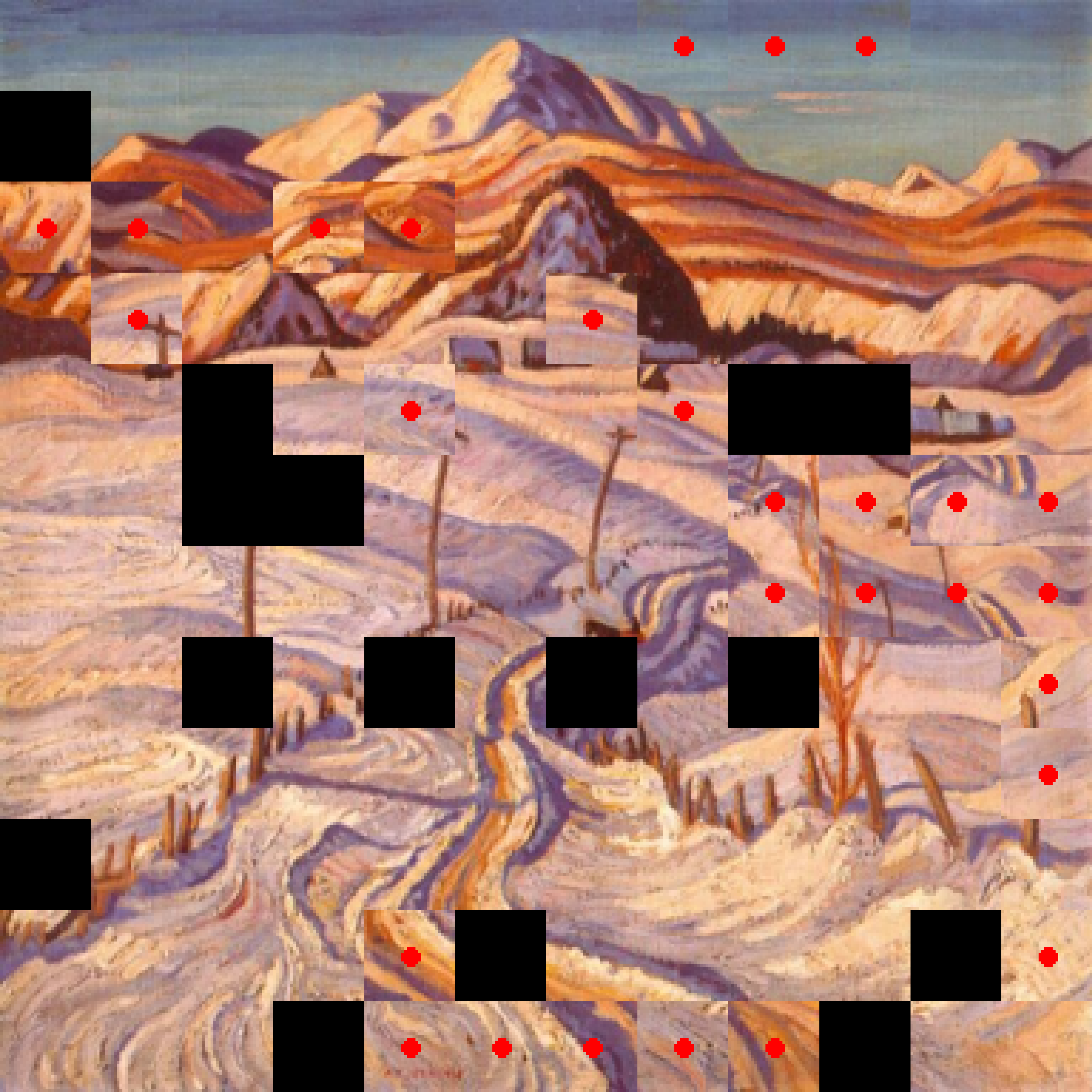}
    & \centering \includegraphics[width=0.16\textwidth]{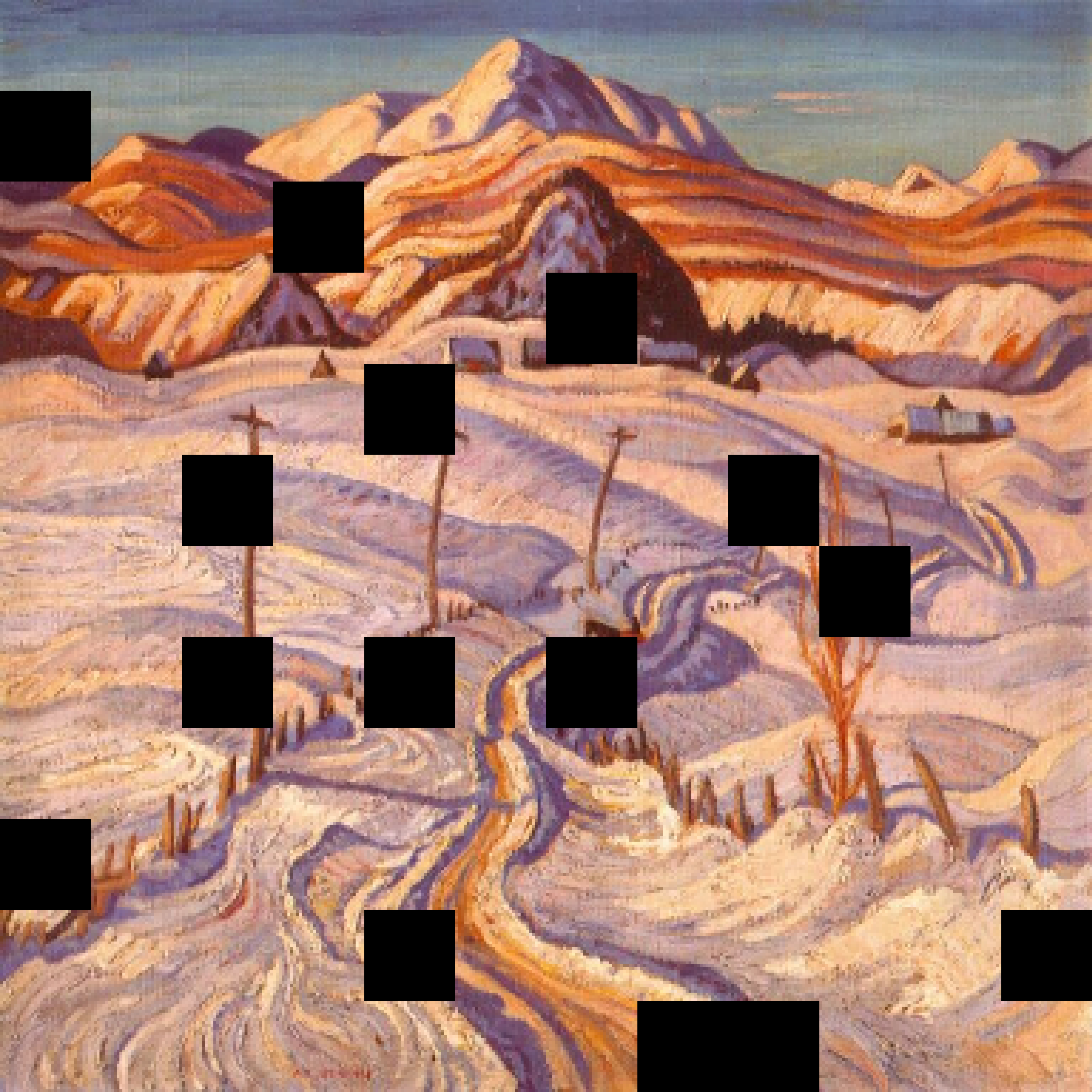} \arraybackslash\\
\centering Eroded\\ Edges\\ (20\%) 
    & \centering \includegraphics[width=0.16\textwidth]{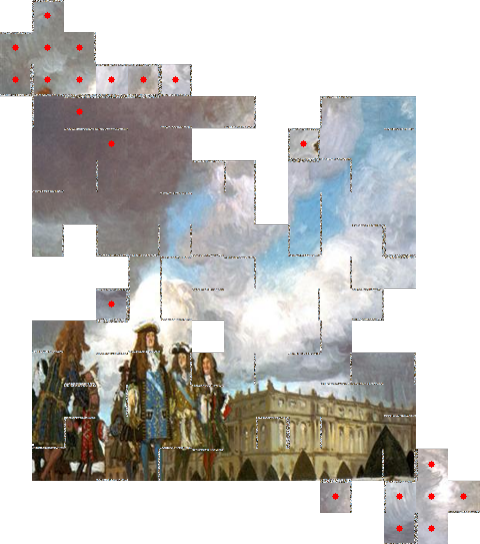} 
    & \centering \includegraphics[width=0.16\textwidth]{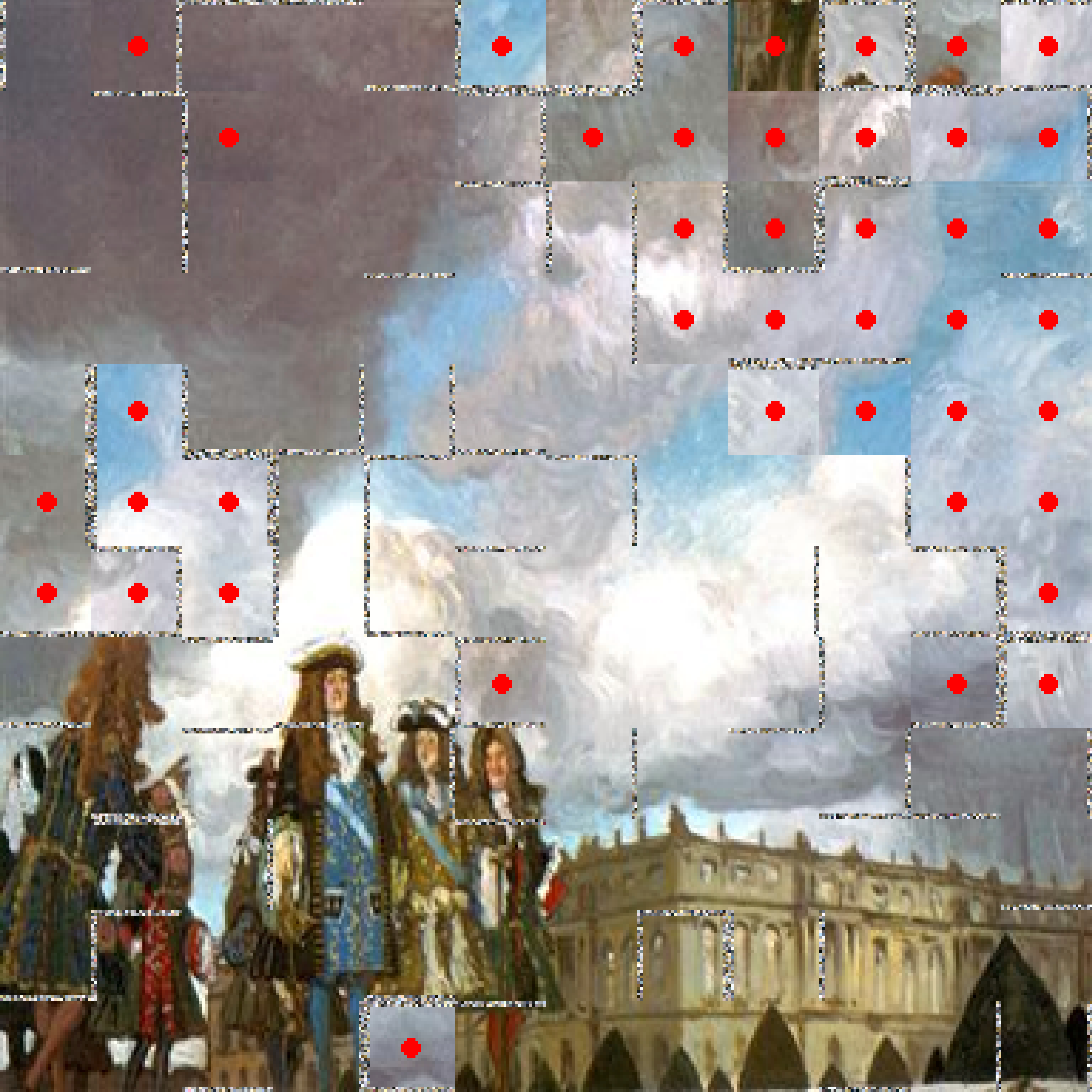} 
    & \centering \includegraphics[width=0.16\textwidth]{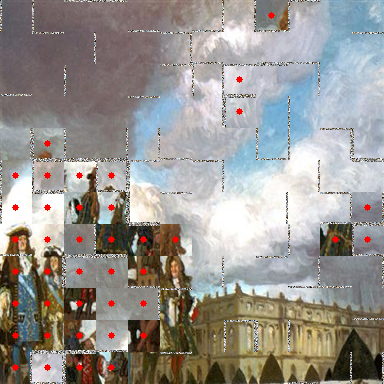} 
    & \centering \includegraphics[width=0.16\textwidth]{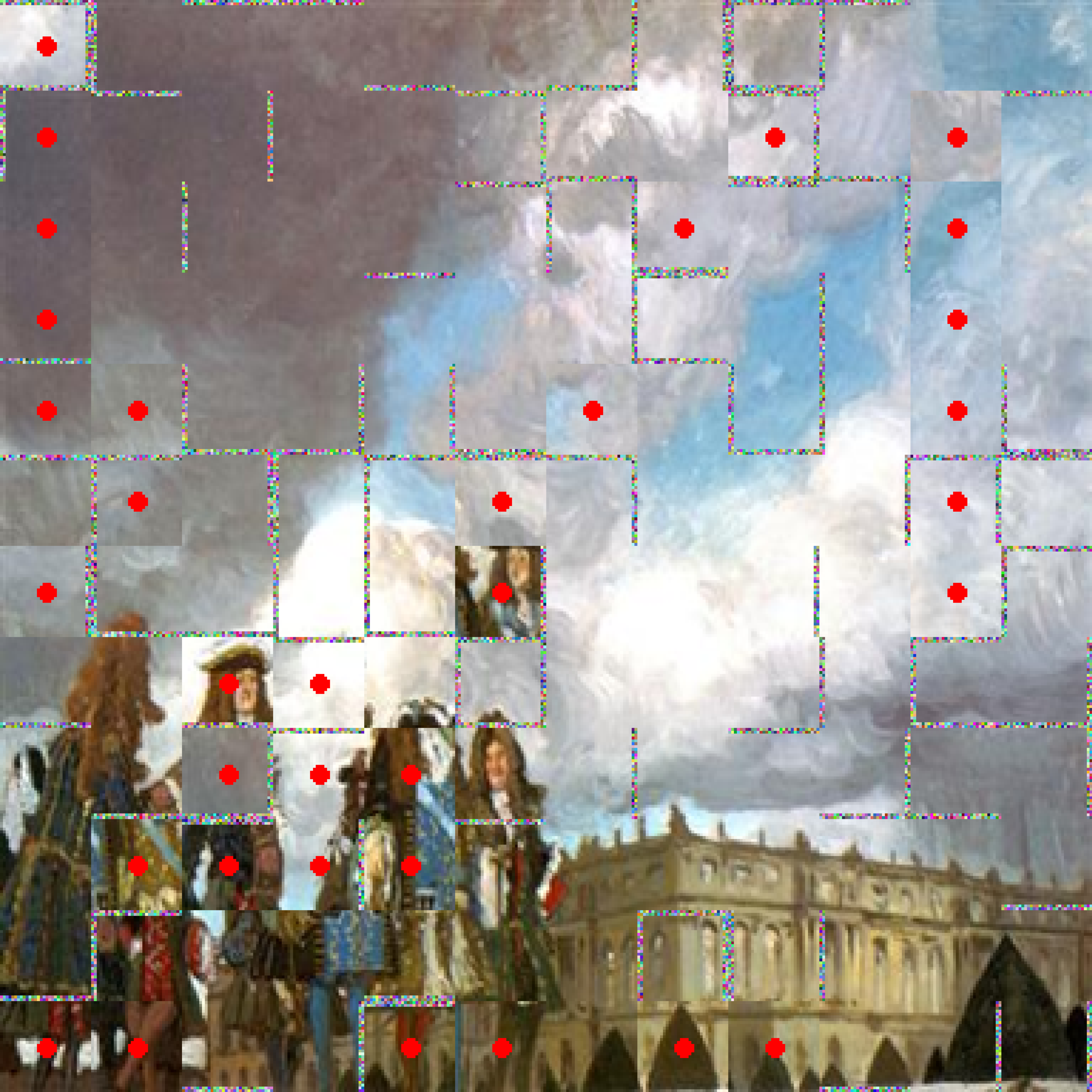} 
    & \centering \includegraphics[width=0.16\textwidth]{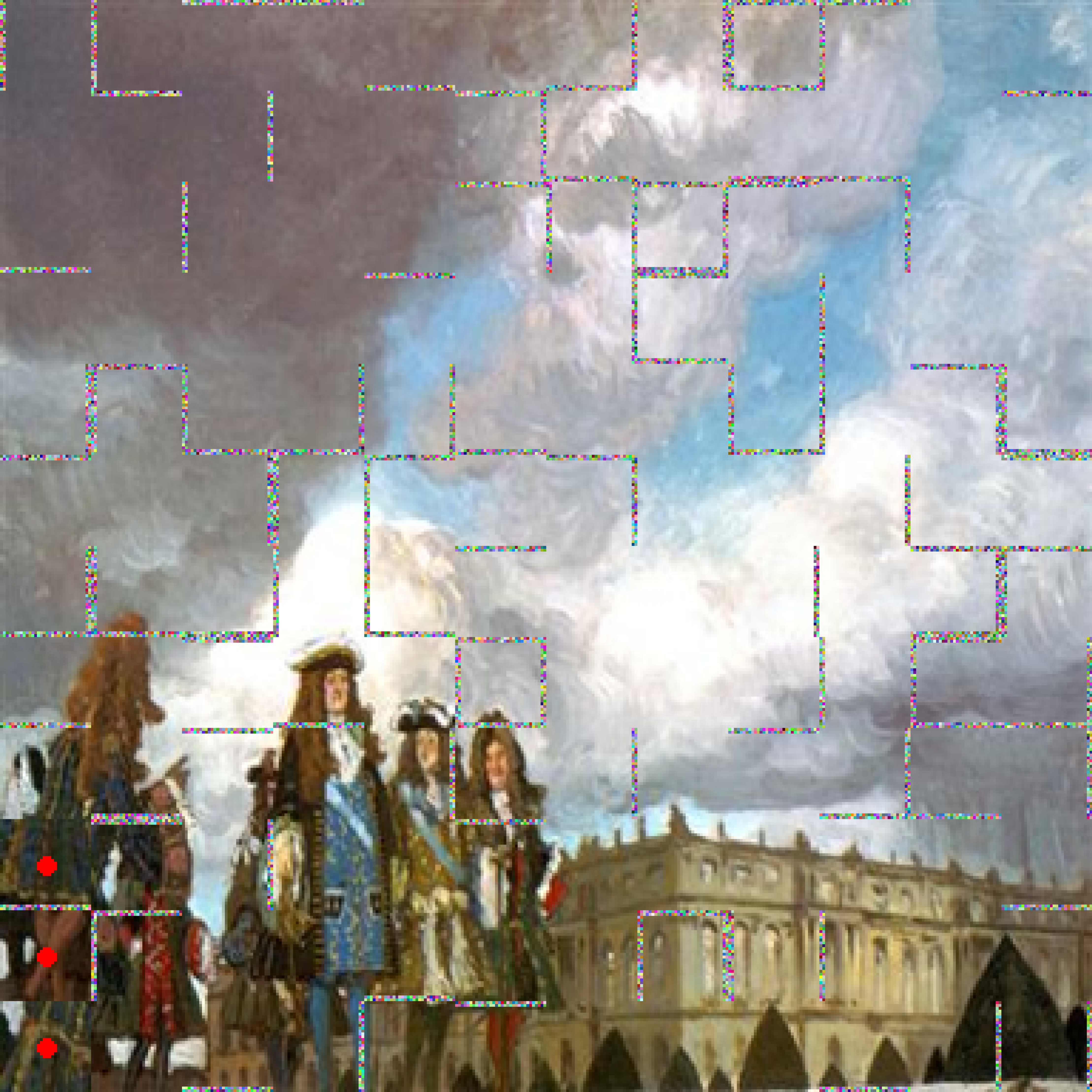} \arraybackslash\\
\centering Eroded\\ Contents\\ (30\%) 
    & \centering \includegraphics[height=0.16\textwidth]{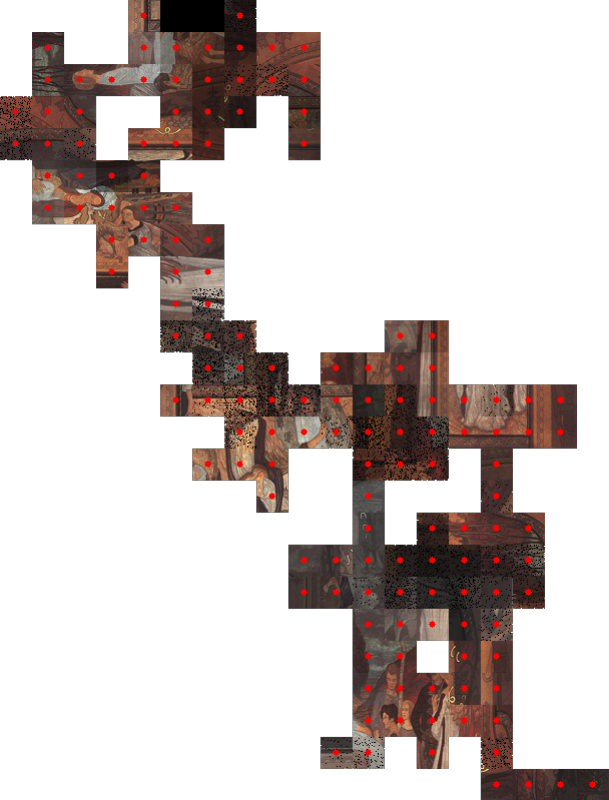} 
    & \centering \includegraphics[width=0.16\textwidth]{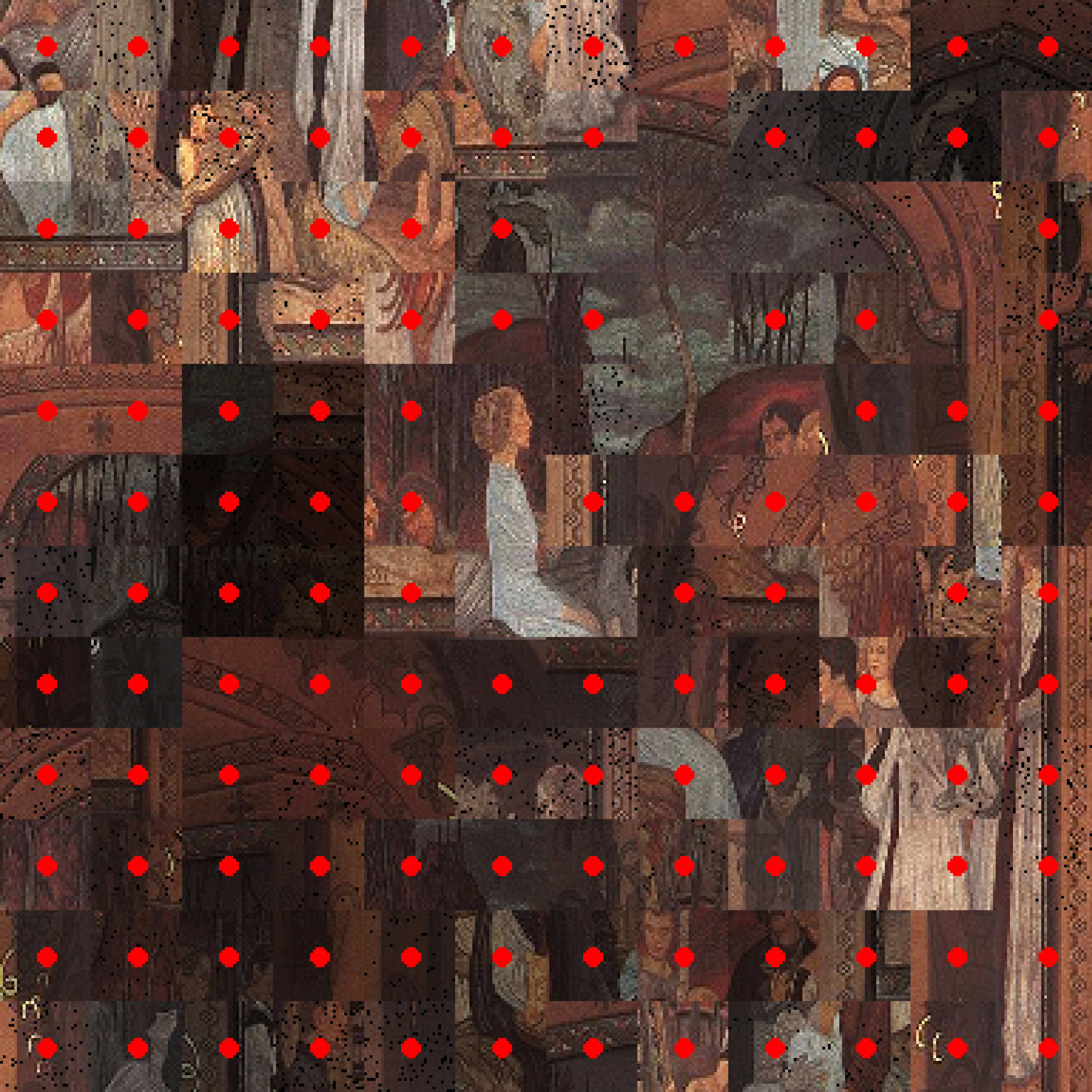} 
    & \centering \includegraphics[width=0.16\textwidth]{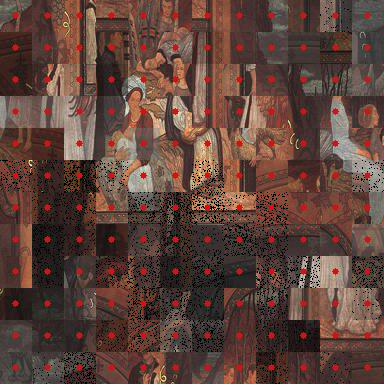} 
    & \centering \includegraphics[width=0.16\textwidth]{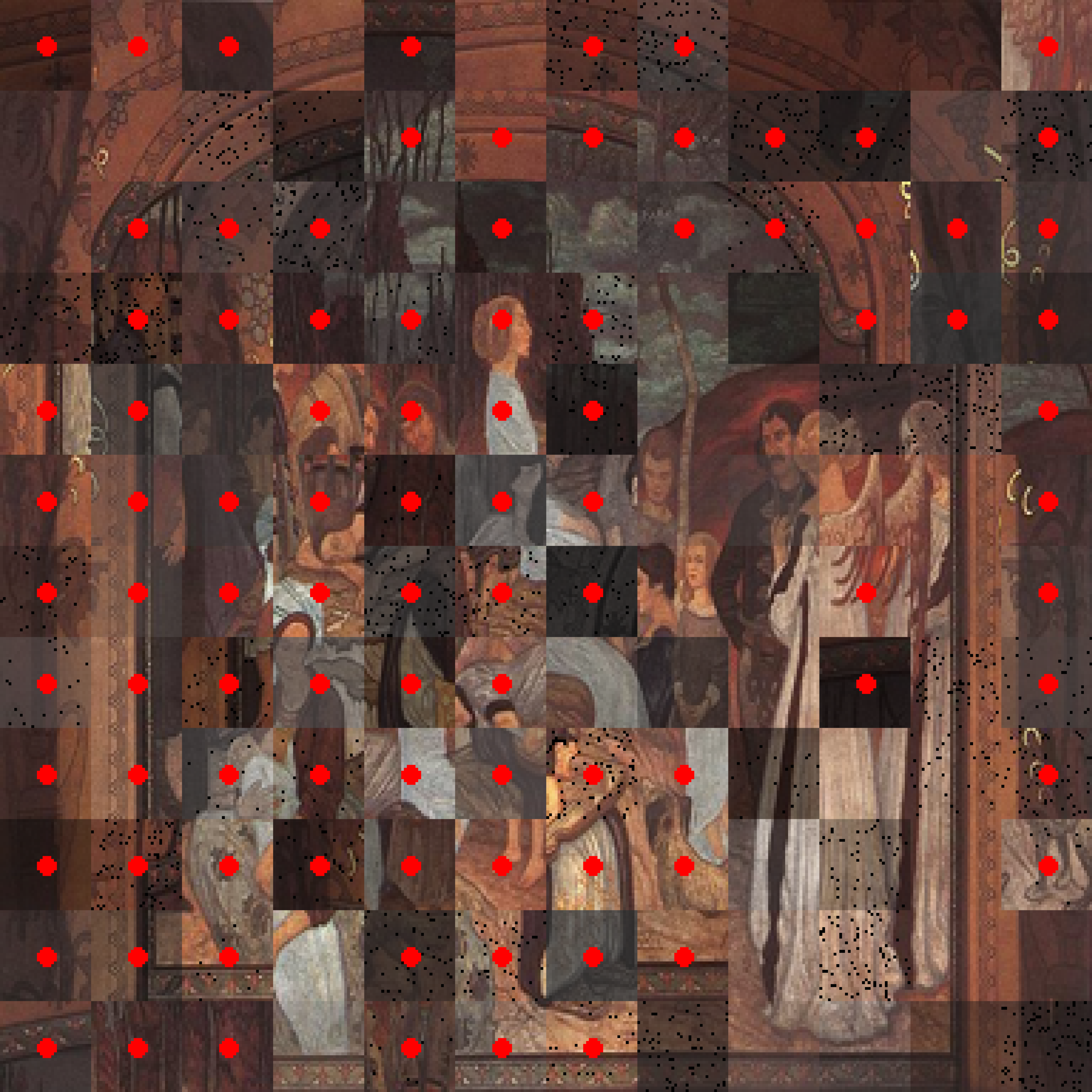} 
    & \centering \includegraphics[width=0.16\textwidth]{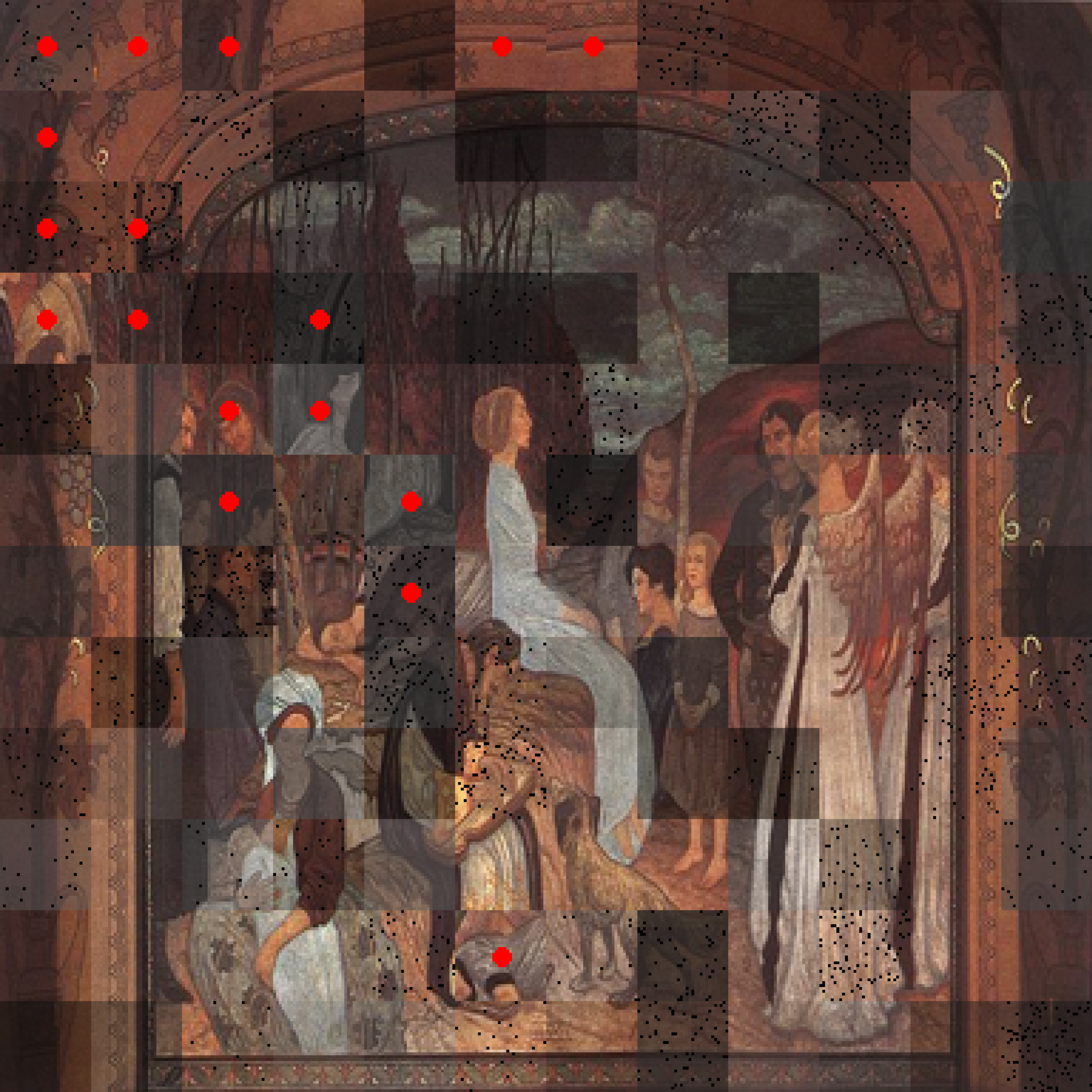} \arraybackslash\\

\bottomrule
\end{tabular}
\end{table}

\section{Discussion}
\label{txt:discussion}

Among the three heuristics, Yu \etal's approach demonstrates the highest robustness across all corruption types. Not only is it the best-performing method on standard Type 1 puzzles, but it also consistently outperforms the approaches by Paikin and Tal and Gallagher across nearly all experiments and evaluation metrics. Since both methods by Gallagher and Yu \etal rely on the same pairwise compatibility metric for identifying neighbouring pieces, this suggests that Yu \etal's linear programming approach is more robust compared to the greedy assembly strategies of the other heuristics. Nevertheless, all three heuristics degrade in performance if more corruption is applied to the puzzles. Corruptions affecting pixel values, such as the eroded edges and eroded contents, lead to a particularly sharp decline, as they directly impact the pairwise compatibility metric of the heuristics. 

Similarly, the base models of the deep learning approaches are not robust to most types of corruption. However, both models demonstrate strong adaptability when trained on augmented data that includes corrupted puzzles. After fine-tuning, their performance improves significantly across all experiments compared to their base versions. The more complex diffusion model consistently outperforms the Transformer model, benefiting from its iterative refinement process during the multiple diffusion steps. 
Fine-tuning particularly enhances the robustness against eroded edges and eroded contents, as the model's embedding layers seem to become less sensitive to the specific type of corruption. Only if pieces are missing, both deep learning models struggle, which results in a lower number of perfectly reconstructed puzzles. Additionally, a general drawback of deep learning models is their declining performance as puzzle size increases and they require large amounts of training data.

\section{Conclusion and Future Work}
\label{txt:conclusion}

This paper explores the robustness and adaptability of content-based jigsaw puzzle solvers when they reassemble corrupted puzzles. Recent research has highlighted that most existing solvers cannot be used to reconstruct real-world fragmented artefacts. As a first step toward bridging this research gap, we benchmarked five SOTA puzzle solvers on three types of corruption, namely missing pieces, eroded edges, and eroded contents. The deep learning models demonstrated their adaptability, with the fine-tuned \textit{Positional Diffusion} model emerging as the most robust solver. It consistently outperformed the heuristic approaches, which rely on pairwise compatibility metrics that become unreliable in the presence of corrupted information. 

Therefore, the development of more advanced deep learning models presents a promising direction for future research. While this study focused on content-based solvers, a key challenge is the combination of content- and shape-based solvers to handle the irregularly shaped fragments that are typically found in real-world artefacts. Additionally, the creation of more realistic synthetic datasets is crucial. These datasets should incorporate more accurate forms of erosion, variable fragment sizes, and irregular shapes to provide better training data for the deep learning models. Another promising direction is the development of hybrid models that combine deep learning with heuristic approaches. Such advancements would bring us closer to fully automated and robust reconstruction methods capable of reassembling fragmented real-world objects.

\bibliographystyle{splncs04}
\bibliography{main}
\end{document}